\documentclass[11pt,a4paper]{article}
\usepackage[hyperref]{acl2021}
\usepackage{times}
\usepackage{latexsym}

\usepackage{microtype}

\aclfinalcopy 
\def\aclpaperid{1980} 

\usepackage{times}
\usepackage{latexsym}
\usepackage{url}
\usepackage{graphicx}
\usepackage{graphics}
\usepackage{multirow}
\usepackage{amsmath}
\usepackage{amssymb}
\usepackage{balance}
\usepackage{mathtools}
\usepackage{varwidth}
\usepackage{subfig}
\usepackage{microtype}
\usepackage{soul}
\usepackage[utf8]{inputenc}
\usepackage{xcolor}
\usepackage{amsthm}
\usepackage{booktabs}
\usepackage{algorithm}
\usepackage{algorithmic}
\usepackage{textcomp}
\usepackage[normalem]{ulem}
\usepackage{array, threeparttable}
\useunder{\uline}{\ul}{}
\usepackage{adjustbox}
\usepackage{dsfont}
\usepackage{bbm}
\usepackage{bm}

\newcommand{\refeqn}[1]{\eqref{#1}}
\newcommand{\reffig}[1]{Figure~\ref{#1}}
\newcommand{\reftbl}[1]{Table~\ref{#1}}
\newcommand{\refsec}[1]{Section~\ref{#1}}
\newcommand{\refapp}[1]{Appendix~\ref{#1}}

\newcommand{\reminder}[1]{}

\newcommand{\etc}{etc.}
\newcommand{\ie}{i.e.,}
\newcommand{\eg}{e.g.,}
\newcommand{\tsne}{t-SNE}
\newcommand{\care}{CaRE}
\newcommand{\conve}{ConvE}
\newcommand{\bert}{BERT}
\newcommand{\bertb}{BERT-base}
\newcommand{\bertl}{BERT-large}
\newcommand{\ufet}{UFET}
\newcommand{\roberta}{RoBERTa}

\newcommand{\system}{OKGIT}

\newcommand{\reverb}{ReVerb}
\newcommand{\reverbk}[1]{ReVerb#1K}
\newcommand{\reverbkf}[1]{ReVerb#1KF}

\newcommand{\sota}{state-of-the-art}

\newcommand{\typescore}{$\psi_{\mathrm{TYPE}}$}
\newcommand{\predscore}{$\psi_{\mathrm{PRED}}$}
\newcommand{\finalscore}{$\psi_{\mathrm{\system{}}}$}
\newcommand{\caretype}{\tau}
\newcommand{\berttype}{\tau_B}

\title{\system{}: Open Knowledge Graph Link Prediction with Implicit Types}

\author{Chandrahas\\
  Indian Institute of Science, Bangalore \\
  \texttt{chandrahas@iisc.ac.in} \\\And
  Partha Pratim Talukdar \\
  Indian Institute of Science, Bangalore \\
  \texttt{ppt@iisc.ac.in} \\}

\date{}

\begin{document}
\maketitle
\begin{abstract}
    Open Knowledge Graphs (OpenKG) refer to a set of (head noun phrase, relation phrase, tail noun phrase) triples such as (tesla, return to, new york) extracted from a corpus using OpenIE tools.
    While OpenKGs are easy to bootstrap for a domain, they are very sparse and far from being directly usable in an end task.
    Therefore, the task of predicting new facts, \ie{} link prediction, becomes an important step while using these graphs in downstream tasks such as text comprehension, question answering, and web search query recommendation.
    Learning embeddings for OpenKGs is one approach for link prediction that has received some attention lately.
    However, on careful examination, we found that current OpenKG link prediction algorithms often predict noun phrases (NPs) with incompatible types for given noun and relation phrases.
    We address this problem in this work and propose \system{} that improves OpenKG link prediction using novel type compatibility score and type regularization.
    With extensive experiments on multiple datasets, we show that the proposed method achieves state-of-the-art performance while producing type compatible NPs in the link prediction task.
\end{abstract}

\section{Introduction}
\label{sec:intro}

An Open Knowledge Graph (OpenKG) is a set of factual triples extracted from a text corpus using Open Information Extraction (OpenIE) tools such as \textsc{TextRunner} \cite{ref:openie} and \reverb{} \cite{ref:reverb}.
These triples are of the form (noun phrase, relation phrase, noun phrase), \eg{} (\textit{tesla}, \textit{return to}, \textit{new york}).
An OpenKG can be viewed as a multi-relational graph where the noun phrases (NPs) are the nodes, and the relation phrases (RPs) are the labeled edges between pairs of nodes.
It is easy to bootstrap OpenKGs from a domain-specific corpus, making them suitable for newer domains.
However, they are extremely sparse and may not be directly usable for an end task.
Therefore, tasks such as NP canonicalization (merging mentions of the same entity) and link prediction (predicting new facts) become an important step in downstream applications.
Some example applications are text comprehension \cite{ref:mausam2016}, relation schema induction \cite{ref:nimishakavi}, canonicalization \cite{ref:cesi}, question answering \cite{ref:qa_freebase}, and web search query recommendation \cite{ref:queryrec}.
In this work, we focus on improving OpenKG link prediction.

\begin{table}[t!]
\centering
\small

    \begin{adjustbox}{max width=0.48\textwidth}
        \begin{tabular}{m{2.5em}m{19.2em}}
             \toprule
             Triple     & (\textit{tesla, return to,\ ?})  \\ 
             \midrule
        \end{tabular}
    \end{adjustbox}
    \begin{adjustbox}{max width=0.49\textwidth}
        \begin{tabular}{m{2.4em}m{3.7em}m{2.4em}m{2.4em}m{3.5em}m{2.0em}}
             \care{}    & \textit{polytechnic institute} & \textit{2009} & \textit{1986} & \textit{jp morgan} & \textit{patent} \\
             \addlinespace[1pt]
             \bert{}    & \textit{chicago} & \textit{earth} & \textit{england} & \textit{america} & \textit{detroit} \\
             \addlinespace[1pt]
             \system{}  & \textit{\underline{new york}} & \textit{america} & \textit{paris} & \textit{california} & \textit{london} \\ 
            \bottomrule
        \end{tabular}
    \end{adjustbox}

    \caption{\label{tab:motivation} Some sample tail NP predictions by \care{}, \bert{}, and \system{}.
        The true tail NP is underlined.
        As we can see, both \care{} and \bert{} fail to predict the correct tail NP.
        However, \bert{} predictions are type compatible with the query.
        \system{} predicts the correct NP while improving the type compatibility with the query.
    }
\end{table}

Although OpenKGs are structurally similar to Ontological KGs, they come with a different set of challenges.
They are extremely sparse, NPs and RPs are not canonicalized, and no type information is present for NPs.
There has been much work on learning embeddings for Ontological KGs in the past years.
However, this task has not received much attention in the context of OpenKGs.
\care{} \cite{ref:care} is a recent method which addresses this problem.
It learns embeddings for NPs and RPs in an OpenKG while incorporating NP canonicalization information.
However, even after incorporating canonicalization, we find that \care{} struggles to predict NPs whose types are compatible with given head NP and RP\@.

As observed by \citet{ref:lama}, modern pre-trained language representation models like \bert{} can store factual knowledge and can be used to perform link prediction in KGs.
However, in our explorations with OpenKGs, we found that even though \bert{} may not predict the correct NP on the top, it predicts type compatible NPs (\reftbl{tab:motivation}).
A similar observation was also made in the context of entity linking \cite{ref:elbert}.
As OpenKGs do not have any underlying ontology and obtaining type information can be expensive, \bert{} predictions can help improve OpenKG link prediction.

Motivated by this, we employ \bert{} for improving OpenKG link prediction, using novel type compatibility score (\refsec{sec:type_score}) and type regularizer term (\refsec{sec:typereg}).
We propose \system{}, a method for OpenKG link prediction with improved type compatibility.
We test our model on multiple datasets and show that it achieves \sota{} performance on all of these datasets.

We make the following contributions:

\begin{itemize}
\setlength\itemsep{0em}
    \item We address the problem of OpenKG link prediction, focusing on improving type compatibility of predictions.
        To the best of our knowledge, this is the first work that addresses this problem.
    \item We propose \system{}, a method for OpenKG link prediction with novel type compatibility score and type regularization.
        \system{} can utilize NP canonicalization information while improving the type compatibility of predictions.
    \item We evaluate \system{} on the link prediction across multiple datasets and observe that it outperforms the baseline methods.
        We also demonstrate that the learned model generates more type compatible predictions.
\end{itemize}

Source code for the proposed model and the experiments from this paper is available at \url{https://github.com/Chandrahasd/OKGIT}.

\begin{figure*}[t]
	\begin{center}
	\includegraphics[scale=0.5]{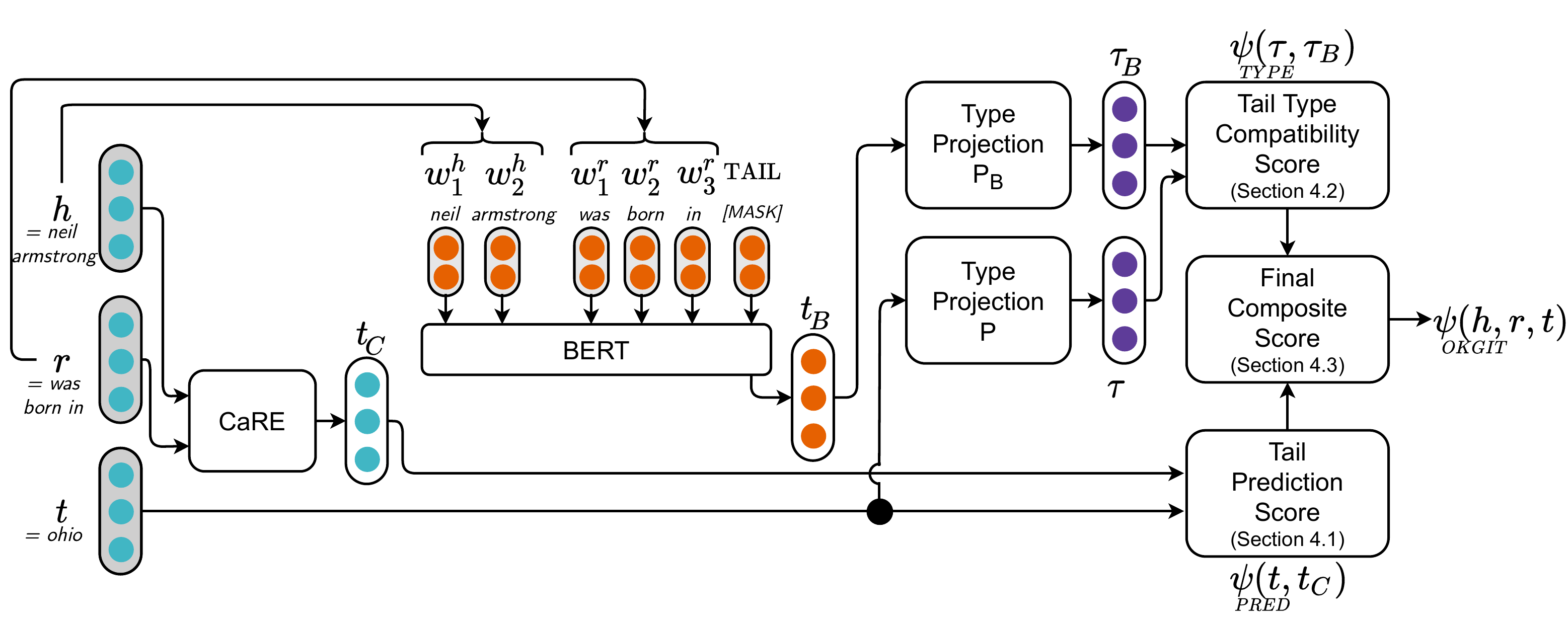}
	\caption{\label{fig:blockdiagram} \system{} Architecture. 
		\system{} learns embeddings for Noun Phrases (NP) and Relation Phrases (RP) present in an OpenKG by augmenting a standard tail prediction loss with type compatibility loss. Guidance for the tail type is obtained through type projection out of BERT's tail embedding prediction.  
		In the figure, $h$, $r$, and $t$ are the head NP, relation (RP), and tail NP.
        $h = w^h_1 \dots w^h_{k_h}$ and $r = w^r_1 \dots w^r_{k_r}$ are tokens in the head NP and relation,  respectively.
		$t_C$ and $t_B$ are the tail NP vectors predicted by \care{} and \bert{} models (Please see \refsec{sec:background} for background on these two models).
        Vectors $\berttype{}$ and $\caretype{}$ are the type vectors obtained using type projections $P_B$ and $P$, respectively.
        \predscore{} represents tail prediction score (\refsec{sec:triple_score}) while \typescore{} represents type compatibility score (\refsec{sec:type_score}).
		\finalscore{} is the combined score generated by \system{} for the input triple $(h,r,t)$ (\refsec{sec:composed_score}).
        Please refer to \refsec{sec:method} for more details.
	}
	\end{center}
\end{figure*}

\section{Related Work}

\noindent \textbf{OpenKG Embeddings}: Learning embeddings for OpenKGs has been a relatively under-explored area of research.
Previous work using OpenKG embeddings has primarily focused on canonicalization.
CESI \cite{ref:cesi} uses KG embedding models for the canonicalization of noun phrases in OpenKGs.
The problem of incorporating canonicalization information into OpenKG embeddings was addressed by \citet{ref:care}.
Their method for OpenKG embeddings (\ie{} \care{}) performs better than Ontological KG embedding baselines in terms of link prediction performance.
The challenges in the link prediction for OpenKGs were discussed in \citet{ref:olpbench}, and methods similar to \care{} were proposed.
In spirit, \care{} \cite{ref:care} comes closest to our model; however, they do not address the problem of type compatibility in the link prediction task.

\noindent \textbf{Entity Type}:
Entity typing is a popular problem where given a sentence and an entity mention, the goal is to predict \emph{explicit} types of the entity.
It has been an active area of research, and many models and datasets, such as \cite{ref:figer}, \cite{ref:ontonotes}, and \cite{ref:ultra}, have been proposed.
However, unlike this task, we aim to incorporate unsupervised \textbf{implicit} type information present in the pre-trained \bert{} model into OpenKG embeddings, rather than predicting \emph{explicit} entity types present in ontologies or corpora.

For unsupervised cases, the problem of type compatibility in link prediction was addressed in \cite{ref:prachitype}.
They employ a type compatibility score by learning a type vector for each NP and two type vectors (head and tail) for each relation.
This score is multiplied with the triple score function, and the type vectors are trained jointly with embedding vectors.
Although their method addresses the type compatibility issue, it is based on Ontological KG embedding models and shares the same limitations.
In another work \cite{ref:tkrl}, hierarchical type information available in the dataset is incorporated while learning embeddings.
However, their model is suitable only for Ontological KGs where the type information is readily available.

\noindent \textbf{\bert{} in KG Embedding}: \bert{} architecture has been used for scoring KG triples \cite{ref:kgbert,ref:coke}.
However, their methods work on Ontological KGs without any explicit attention to NP types.
In other work \cite{ref:lama}, pre-trained \bert{} models are used for predicting links in KG\@.
However, their focus was to evaluate knowledge present in the pre-trained \bert{} models instead of improving the existing link prediction model.
\bert{} embeddings were also used for extracting entity type information \cite{ref:elbert}.
However, it was used for Entity Linking compared to OpenKG link prediction in our case.
\section{Background}
\label{sec:background}

We first introduce the notation used in this paper, followed by brief descriptions of \bert{} and \care{}.

\noindent \textbf{Notation}:
An Open Knowledge Graph OpenKG = ($\mathcal{N}$, $\mathcal{R}$, $\mathcal{T}$) contains a set of noun phrases (NPs) $\mathcal{N}$, a set of relation phrases (RPs) $\mathcal{R}$ and a set of triples $(h, r, t) \in \mathcal{T}$ where $ h,t\in \mathcal{N}$ and $r \in \mathcal{R}$.
Here, $h$ and $t$ are called the head and tail NPs, and $r$ is the RP between them.
Each of them contains tokens from a vocabulary $\mathcal{V}$, specifically, $h = (w_1^h, w_2^h, \dots, w_{k_h}^h)$, $t = (w_1^t, w_2^t, \dots, w_{k_t}^t)$ and $r = (w_1^r, w_2^r, \dots, w_{k_r}^r)$. 
Here, $k_h$, $k_r$, and $k_t$ are the numbers of tokens in the head NP, the relation, and the tail NP.
OpenKG embedding methods learn vector representations for NPs and RPs.
Specifically, vectors for an NP $e\in \mathcal{N}$ and an RP $r\in \mathcal{R}$ are represented by boldface letters $\mathbf{e}\in \mathbb{R}^{d_e}$ and $\mathbf{r}\in \mathbb{R}^{d_r}$.
Here, $d_e$ and $d_r$ are dimensions of NP and RP vectors. Usually, $d_e = d_r$.
A score function $\psi(h, r, t)$ represents the plausibility of a triple.
Similarly, \bert{} represents tokens by $d_B$-dimensional vectors.
A type projection matrix $P$ takes the vectors to a common $d_{\tau}$-dimensional type space $\mathbb{R}^{d_{\tau}}$.
The vectors in the type space are denoted by $\mathbf{\tau}$.

\noindent \textbf{\bert{} \cite{ref:bert}}: \bert{} is a bi-directional language representation model based on the transformer architecture \cite{ref:transformer}, which has shown performance improvements across multiple NLP tasks.
It is pre-trained on two tasks,
(1) Masked Language Modeling (MLM), where the model is trained to predict randomly masked tokens from the input sentences, and
(2) Next Sentence Prediction (NSP), where the model is trained to predict whether an input pair of sentences occurs in a sequence or not.
In our case, we use a pre-trained \bert{} model (without fine-tuning) for predicting a masked tail NP in a triple.

\noindent \textbf{\care{} \cite{ref:care}}: \care{} is an OpenKG embedding method that can incorporate NP canonicalization information while learning the embeddings.
NP canonicalization is the problem of grouping all surface forms of a given entity in one cluster, e.g., inferring that \textit{Barack Obama}, \textit{Barack H. Obama}, and \textit{President Obama} all refer to the same underlying entity.
\care{} consists of three components: 
(1) a canonicalization cluster encoder (CN), which generates NP embeddings by aggregating embeddings of canonical NPs from the corresponding cluster,
(2) a bi-directional GRU based phrase encoder (PN), which encodes the tokens in RPs to generate RP embeddings, and 
(3) a base model, which is an Ontological KG embedding method like \conve{} \cite{ref:conve}. It uses NP and RP embeddings for scoring triples.
These triple scores are then fed to a loss function (\eg{} pairwise ranking loss with negative sampling \cite{ref:transE} or binary cross-entropy loss (BCE) \cite{ref:conve}).
In this paper, we use \care{} with \conve{} as the base model.
This model generates a candidate tail NP vector for a given NP $h$ and RP $r$, denoted by \care{}($\mathbf{h}$, $\mathbf{r}$).

\section{\system{}: Our Proposed Method}
\label{sec:method}

\noindent \textbf{Motivation}: As illustrated in \reftbl{tab:motivation}, top NPs predicted by \care{} may not always be type compatible with the input query.
On the other hand, \bert{}'s top predictions are usually type compatible \cite{ref:elbert}, although they may not be factually correct.
Thus, we hypothesize that a combination of these two models can produce correct as well as type compatible predictions.
Motivated by this, we develop \system{}, which combines the best of both of these models.
The complete architecture of the proposed model can be found in \reffig{fig:blockdiagram}.
In the following section, we present various components of the proposed model.

\subsection{\predscore{}: Tail Prediction Score}
\label{sec:triple_score}

The correctness of tail prediction in a triple is measured by the triple score function \predscore{}.
Given a triple $(h,r,t)$, it uses the corresponding vectors ($\mathbf{h}$, $\mathbf{r}$, $\mathbf{t}$) and assigns high scores to correct triples and low scores to incorrect triples.
We follow \care{} \cite{ref:care} for scoring triples, which internally uses \conve{} \cite{ref:conve} as the base model.
For a given triple $(h,r,t)$, the \care{} model first predicts a tail NP vector $\mathbf{t_C}$ as
\begin{equation} \label{eqn:carevec}
    \mathbf{t_C} = \text{\care{}}(\mathbf{h},\mathbf{r})
\end{equation}
The predicted tail NP vector $\mathbf{t_C}$ is then matched against the given tail NP vector $\mathbf{t}$ using dot product to generate the triple score \predscore{}.
\begin{equation}
    \psi_{\mathrm{PRED}}(\mathbf{t}, \mathbf{t_C}) = \mathbf{t_C}^\top \mathbf{t}.
\end{equation}

The score \predscore{} represents tail prediction correctness, and \care{} model uses only this score.

\subsection{\typescore{}: Tail Type Compatibility Score}
\label{sec:type_score}

The type compatibility between a given (head NP, RP) pair and a tail NP is measured by the type compatibility score function \typescore{}.
It assigns a high score when an NP $t$ has suitable types as candidate tail NP for given head NP $h$ and RP $r$.
We employ a Masked Language Model (MLM) for measuring type compatibility, specifically \bert{} \cite{ref:bert}.
Following \cite{ref:lama}, we can generate a candidate tail NP vector using \bert{}.
Specifically, given a triple $(h,r,t)$, we replace the head NP $h$ and RP $r$ with their tokens and tail NP $t$ with a special MASK token.
The resulting sentence $(w_1^h, \dots, w_{k_h}^h, w_1^r, \dots, w_{k_r}^r, \text{MASK})$ is sent as input to the \bert{} model. 
We denote the output vector from \bert{} corresponding to the MASK tail token as $\mathbf{t_B}$. 
\begin{equation}\label{eqn:bertvec}
    \mathbf{t_B} = \text{\bert{}}(h, r, \text{MASK}) 
\end{equation}

We can predict tail NPs for a given (h, r) by finding the nearest neighbors of $\mathbf{t_B}$ from the \bert{} vocabulary (\refapp{sec:bertAsLPM}).
These predicted NPs may not be the correct tail NP present in KG; however, they tend to be type compatible with the given ($h$, $r$) pair.

Motivated by this, we extract the implicit NP type information from this vector using a type projector $P_B \in \mathbb{R}^{d_{\tau} \times d_B}$.
The output vector from \bert{} $t_B$ is high-dimensional and can be used as a proxy for NP's type \cite{ref:elbert}.
Therefore, $P_B$ projects the $\mathbf{t_B}$ vector to a lower dimensional space such that only relevant information is retained.
We do a similar operation on tail NP embedding $\mathbf{t}$ and use a type projector $P \in \mathbb{R}^{d_{\tau} \times d_e}$ to extract type information.
Both $P_B$ and $P$ are trained jointly with the model.
Thus, the type vectors are given by
\begin{equation} \label{eqn:typevec}
	\bm{\berttype{}} = P_B \mathbf{t_B} \hspace{3mm}  \text{and} \hspace{3mm} \bm{\caretype{}} = P \mathbf{t}
\end{equation}
for \bert{} and \care{}, respectively.
Here, both $\bm{\berttype{}}, \bm{\caretype{}} \in \mathbb{R}^{d_{\tau}}$.
Then, the type compatibility score between these can be measured by negative of Euclidean distance, \ie{}
\begin{equation*}
    \psi_{\mathrm{TYPE}}(\bm{\caretype{}}, \bm{\berttype{}}) = -||\bm{\berttype{}} - \bm{\caretype{}}||^2_2.
\end{equation*}

We also experimented with a dot product version of the type score,  $\psi^{\mathrm{Dot}}_{\mathrm{TYPE}}(\bm{\caretype{}}, \bm{\berttype{}}) = \bm{\berttype{}}^\top \bm{\caretype{}}$, and found its performance to be comparable to the Euclidean distance version.
Therefore, we use the Euclidean distance version for all our experiments.

\subsection{\finalscore{}: Final Composite Score}
\label{sec:composed_score}

The score functions \predscore{} and \typescore{} may contain complementary information.
Therefore, we use a combination of triple and type compatibility scores as final score for a given triple.
\begin{equation} \label{eqn:scoreokget}
    \resizebox{0.49\textwidth}{!}{\
        $\psi_{\mathrm{\system{}}}(h, r, t) = \psi_{\mathrm{PRED}}(\mathbf{t}, \mathbf{t_C}) + \gamma \times \psi_{\mathrm{TYPE}}(\bm{\caretype{}}, \bm{\berttype{}})$.
    }
\end{equation}

Please recall that $\mathbf{t_C}$ and $\bm{\berttype{}}$ are in turn dependent on
$h$ and $r$ (\refeqn{eqn:carevec} and \refeqn{eqn:bertvec}), while $\bm{\caretype{}}$ is dependent on $t$ (\refeq{eqn:typevec}).
Here, $\gamma$ controls the relative weights given to individual scores.
This final score takes care of both, \ie{} triple correctness as well as type compatibility.
For training, we feed the sigmoid of this score function to the Binary Cross Entropy (BCE) loss function following \cite{ref:conve}.

\subsection{Learning with Type Regularization}
\label{sec:typereg}

Let $\mathcal{X} = \{(h_i, r_i)| (h_i, r_i, t_i) \in \mathcal{T} \text{for some } t_i \in \mathcal{N}\}$ be the set of all head NPs and RPs which appear in the OpenKG\@.
Let $y_{i}$ be the label for the triple $(h_i, r_i, t_i)$ which is $1$ if $(h_i, r_i, t_i) \in \mathcal{T}$ and $0$ otherwise.
We apply the logistic sigmoid function $\sigma$ on score \finalscore{} to get the predicted label 
\[
	\hat{y}_{i} = \sigma(\psi_{\mathrm{\system{}}}(h_i, r_i, t_i))
\]
Finally, we use the following binary cross-entropy (BCE) loss for triple correctness.
\begin{equation*}
   \resizebox{0.49\textwidth}{!}{\
        $\text{TripleLoss}(h_i, r_i, t_i) = y_{i}\cdot{}\log(\hat{y}_{i}) + (1-y_{i})\cdot{}\log(1-\hat{y}_{i})$
    }
\end{equation*}

\begin{table}[t]
    \begin{adjustbox}{max width=0.47\textwidth}
    \begin{tabular}{m{6em}m{3em}m{3em}m{3em}m{5em}}
        \toprule
            Dataset       & \#NPs   & \#RPs   & \begin{tabular}{@{}c@{}}\#Gold \\ Clusters \end{tabular} & \begin{tabular}{@{}c@{}}\#Average NPs\\ Per Cluster \end{tabular} \\
        \midrule
            \reverbk{20}  & 11,064     & 11,057     & 10,897    & \multicolumn{1}{c}{1.02} \\
            \reverbk{45}  & 27,007     & 21,622     & 18,626    & \multicolumn{1}{c}{1.45} \\
            \reverbkf{20} & \ \ 3,524  & \ \ 6,076  & \ \ 3,406 & \multicolumn{1}{c}{1.03} \\
            \reverbkf{45} & \ \ 9,400  & 11,249     & \ \ 6,749 & \multicolumn{1}{c}{1.39} \\ 
    \end{tabular}
     \end{adjustbox}
    \begin{adjustbox}{max width=0.47\textwidth}
    \begin{tabular}{m{6em}m{5.3em}m{5em}m{3.7em}}
        \toprule
            & \#Train   & \#Validation  & \#Test \\
        \midrule
            \reverbk{20}  & 15,498      & 1,549 & 2,324   \\
            \reverbk{45}  & 35,969      & 3,597 & 5,394   \\
            \reverbkf{20} & \ \ 6,685   & 1,015 & 1,517   \\
            \reverbkf{45} & 14,775      & 1,781 & 2,650   \\ 
        \bottomrule
    \end{tabular}
     \end{adjustbox}
    \caption{\label{tab:datasets}Dataset Statistics. Please refer to {\refsec{sec:datasets}} for more details.}
\end{table}

To further reinforce the type compatibility in the model, we include an additional loss term which forces the type vectors of correct triples to be closer in the type space.
Similar to $\text{TripleLoss}$, we use the binary cross-entropy loss for type regularization as well.
The type regularization term is shown below.
\begin{equation*}
    \resizebox{0.49\textwidth}{!}{\
        $\text{TypeLoss}(h_i, r_i, t_i) = y_{i} \cdot{} \log(\hat{p}_{i}) + (1-y_{i})\cdot{}\log(1-\hat{p}_{i})$
    }
\end{equation*}
\noindent where $\hat{p} = \sigma(\psi_{\mathrm{TYPE}}(\bm{\caretype{}}, \bm{\berttype{}}))$. The cumulative loss function is then given as below.
\begin{equation} \label{eqn:totalloss}
	\sum_{i = 1}^{n} \text{TripleLoss}(h_i, r_i, t_i) + \lambda \times \text{TypeLoss}(h_i, r_i, t_i)
\end{equation}
where $n$ is the number of training instances.
We consider $\mathcal{X} \times \mathcal{N}$ as our training data where triples present in $\mathcal{T}$ have label 1 and rest have label 0.

\section{Experiments}

\label{sec:datasets}
\noindent \textbf{Datasets}:
Following \cite{ref:care}, we use two subsets of English OpenKGs created using \reverb{} \cite{ref:reverb}, namely \reverbk{20} and \reverbk{45}.
We follow the same train-validation-test split for these datasets.
As noted in \cite{ref:lama}, predicting multi-token NPs using \bert{} could be challenging and it might require special pre-training \cite{ref:spanbert}.
To understand this difference, we create filtered subsets of these datasets such that they contain only single token NPs
\footnote{Please note that the single-token limitation is only valid for \bert{}, not for \system{} (\refapp{sec:bertAsLPM}).}.
Specifically, we create \reverbkf{20} (\reverbk{20}-Filtered) and \reverbkf{45} (\reverbk{45}-Filtered) which contain only single token NPs.
More details about these datasets can be found in \reftbl{tab:datasets}.

\begin{table}[t!]
    \begin{adjustbox}{max width=0.48\textwidth}
    \begin{tabular}{m{6em}m{3.5em}m{2.5em}m{2.5em}m{2.5em}m{3em}}
        \toprule
        Dataset         & $d_e = d_r$   & \multicolumn{1}{c}{$d_{\tau}$} & \multicolumn{1}{c}{$\lambda$}\  & \multicolumn{1}{c}{$\gamma$}\  & \bert{} model \\
        \midrule
        \reverbk{20}    & \ \ \ 300     & 300        & 0.01      & 5.0      & large \\
        \reverbk{45}    & \ \ \ 300     & 100        & 0.0       & 2.0      & large \\
        \reverbkf{20}   & \ \ \ 300     & 300        & 0.001     & 5.0      & base \\
        \reverbkf{45}   & \ \ \ 300     & 300        & 0.001     & 0.25     & base \\
        \bottomrule
    \end{tabular}
    \end{adjustbox}
    \caption{\label{tab:hyperparams}Optimal Hyperparameter values. Please refer to {\refsec{sec:hyperparams}} for more details.}
\end{table}
\label{sec:hyperparams}
\noindent\textbf{Setup and hyperparameters}: We use $d_e$ = $d_r$  = $300$ for NP and RP vectors.
For other hyper-parameters, we use grid-search and select the model based on MRR on validation split.
For type vectors, we select $d_{\tau}$ from \{100, 300, 500\}.
The weight for type regularization term $\lambda$ is selected from the range $\{10^{-3}, 10^{-2}\dots, 10^1\} \cup \{0\}$.
Type composition weight $\gamma$ is selected from \{$0.25$, $0.5$, $1.0$, $2.0$, $5.0$\}.
For the language model, we try both \bertb{} as well as \bertl{}.
The optimal values for hyperparameters are shown in \reftbl{tab:hyperparams}.
The experiments run for 1.5 hours (for filtered subsets) and 9 hours (for full datasets) on GeForce GTX 1080 Ti GPU.

\begin{table*}[thb]
\centering
\begin{adjustbox}{max width=\textwidth}
\begin{tabular}{m{12em}m{4em}m{3em}m{3em}m{3em}m{3em}m{0em}m{4em}m{3em}m{3em}m{3em}m{3em}}
\toprule
\textbf{}         & \multicolumn{5}{c}{\textbf{\reverbk{20}}}                         & & \multicolumn{5}{c}{\textbf{\reverbk{45}}}                            \\ \cmidrule{2-6} \cmidrule{8-12}
\textbf{Model}    & \textbf{MRR}($\%$)$\uparrow$ & \textbf{MR}$\downarrow$ & \multicolumn{3}{c}{\textbf{Hits}($\%$)$\uparrow$} & & \textbf{MRR}($\%$)$\uparrow$ & \textbf{MR}$\downarrow$ & \multicolumn{3}{c}{\textbf{Hits}($\%$)$\uparrow$} \\ \cmidrule{4-6} \cmidrule{10-12}
                  &             &              & {\bf @1}         & {\bf @3}        & {\bf @10}       & &              &              & {\bf @1}         & {\bf @3}        & {\bf @10}       \\ \toprule
{\conve{} \cite{ref:conve} }    & 26.2  & 2177.0  & 20.2   & 29.1  & 36.3  & & 18.4  & 6625.0  & 13.3  & 20.6  & 28.3 \\ 
{\care{} \cite{ref:care} }      & 30.6  & 851.1   & 24.4   & 33.1  & 41.7  & & 32.0  & 1276.8  & 25.3  & 35.0  & 44.6 \\ 
{\care{} [BERT initialization]} & 31.6	& 837.0   & 24.8   & 35.0  & 44.2  & & 31.2	 & 925.5   & 24.2  & 34.4  & 44.3 \\
{\system{} [Our model]}         & \textbf{35.9} & \textbf{527.1}  & \textbf{28.2} & \textbf{39.4} & \textbf{49.9} & & \textbf{33.2} & \textbf{773.9}  & \textbf{26.1} & \textbf{36.3} & \textbf{46.4} \\
\bottomrule
\end{tabular}
\end{adjustbox}
\begin{adjustbox}{max width=\textwidth}
\begin{tabular}{m{12em}m{4em}m{3em}m{3em}m{3em}m{3em}m{0em}m{4em}m{3em}m{3em}m{3em}m{3em}}
\toprule
\textbf{} & \multicolumn{5}{c}{\textbf{\reverbkf{20}}} & & \multicolumn{5}{c}{\textbf{\reverbkf{45}}} \\ \cmidrule{2-6} \cmidrule{8-12}
\textbf{Model} & \textbf{MRR}($\%$)$\uparrow$ & \textbf{MR}$\downarrow$ & \multicolumn{3}{c}{\textbf{Hits}($\%$)$\uparrow$} & & \textbf{MRR}($\%$)$\uparrow$ & \textbf{MR}$\downarrow$ & \multicolumn{3}{c}{\textbf{Hits}($\%$)$\uparrow$} \\ \cmidrule{4-6} \cmidrule{10-12}
                  &             &              & {\bf @1}         & {\bf @3}         & {\bf @10}       & &              &              & {\bf @1}         & {\bf @3}        & {\bf @10}       \\ \toprule

{\bert{} \cite{ref:bert} }      & 4.9  & 1116.5 & 2.2  & 5.0  & 9.7  & & 18.9 & 536.5  & 12.3 & 20.8 & 32.5 \\ 
{\conve{} \cite{ref:conve} }    & 22.3 & 836.6  & 16.1 & 25.5 & 33.4 & & 16.5 & 2398.1 & 10.9 & 18.9 & 27.6 \\ 
{\care{} \cite{ref:care}}       & 29.3 & 308.3  & 22.1 & 31.6 & 43.2 & & 26.6 & 692.7  & 20.1 & 28.8 & 39.1 \\
{\care{} [BERT initialization]} & 31.8 & \textbf{207.6}	& 24.2 & 34.8 & 46.2 & & 24.9 & 557.3  & 17.8 & 27.6 & 38.3 \\
{\system{} [Our model]}         & \textbf{34.6} & 214.7  & \textbf{26.5} & \textbf{38.0} & \textbf{50.2} & & \textbf{29.7} & \textbf{500.2}  & \textbf{22.5} & \textbf{32.4} & \textbf{43.3} \\
\bottomrule
\end{tabular}
\end{adjustbox}
\caption{\label{tab:lpresults} Results of link prediction task.
        Here $\uparrow$ indicates higher values are better while $\downarrow$ indicates lower values are better.
        We can see that the \system{} model outperforms the baseline models on all the datasets (\refsec{sec:lpresults}).
    }
\end{table*}

\section{Results}
We evaluate the proposed model on the link prediction task.
We follow the same evaluation process as in \cite{ref:care}.
From our experiments, we try to answer the following questions:
\begin{enumerate}
	\item Is \system{} effective in the link prediction task? (\refsec{sec:lpresults})
    \item Does \system{} generate more type compatible NPs in link prediction? (\refsec{sec:okgetTypeEval})
    \item Is the Type Projector effective in extracting type vectors from embeddings? (\refsec{sec:type_vec})
\end{enumerate}

\subsection{Effectiveness of \system{} Embeddings in Link Prediction}
\label{sec:lpresults}

We evaluate our model on the link prediction task.
Given a held-out triple $(h_i,r_i,t_i)$, all the NPs $e\in\mathcal{N}$ in the KG are ranked as candidate tail NP based on their score $\psi_{\mathrm{\system{}}}(h_i,r_i,e)$.
Let the rank of the correct tail NP $t$ be denoted by $rank^t_i$.
Similarly, ranks are also calculated for predicting head NPs instead of tail NPs using inverse relations \cite{ref:conve, ref:care}; let it be denoted by $rank^h_i$.
These ranks are then used to find Mean Reciprocal Rank (MRR), Mean Rank (MR) and Hit@k (k=1,3,10) as follows.

\begin{equation*}
    \text{MRR} = \frac{1}{2\times n_{test}}\sum_{i = 1}^{n_{test}} \left(\frac{1}{rank^h_i} + \frac{1}{rank^t_i}\right),
\end{equation*}

\begin{equation*}
    \text{MR} = \frac{1}{2\times n_{test}}\sum_{i = 1}^{n_{test}} \left(rank^h_i + rank^t_i\right), \text{and}
\end{equation*}

\begin{equation*}
    \text{Hits@k} = \sum_{i = 1}^{n_{test}} \frac{\mathbbm{1}(rank^h_i \leq k) + \mathbbm{1}(rank^t_i \leq k)}{2\times n_{test}}.
\end{equation*}

Here, $n_{test}$ is the number of test triples and $\mathbbm{1}$ is the indicator function.
As noted in \cite{ref:care}, ranking individual NPs is not suitable for OpenKGs due to the lack of canonicalization.
Hence, following their approach, we rank gold canonicalization clusters instead of individual NPs.
The gold canonicalization partitions the NPs into clusters such that NPs mentioning the same entity belong to the same cluster.
For ranking these clusters, we first find ranks of all NPs $e \in \mathcal{N}$. Then for each cluster, we keep the NP with minimum rank as representative and discard others. The representative NPs are then ranked again and the new ranks are assigned to the corresponding clusters. The rank of the cluster containing the true NP is then used for evaluating the performance.
For better readability, the MRR and Hits@k metrics have been multiplied by 100.

We compare \system{} with \bert{} (MLM), \conve{} (Ontological KGE) and \care{} (OpenKGE).
We also compare against a version of \care{} where phrase embeddings have been initialized with \bert{} (\care{} [\bert{} initialization]).
As we can see from the results in \reftbl{tab:lpresults}, the proposed model \system{} outperforms baseline methods in link prediction task across all datasets.
This suggests that the implicit type scores from \bert{} help in improving ranks of correct NPs.
Moreover, \system{} outperforms \care{} with \bert{} initialization, suggesting the importance of type projectors
\footnote{Please refer to \refapp{sec:baselines} for a detailed comparison.}.

\begin{figure}[thb]
    \begin{center}
        \includegraphics[scale=0.22]{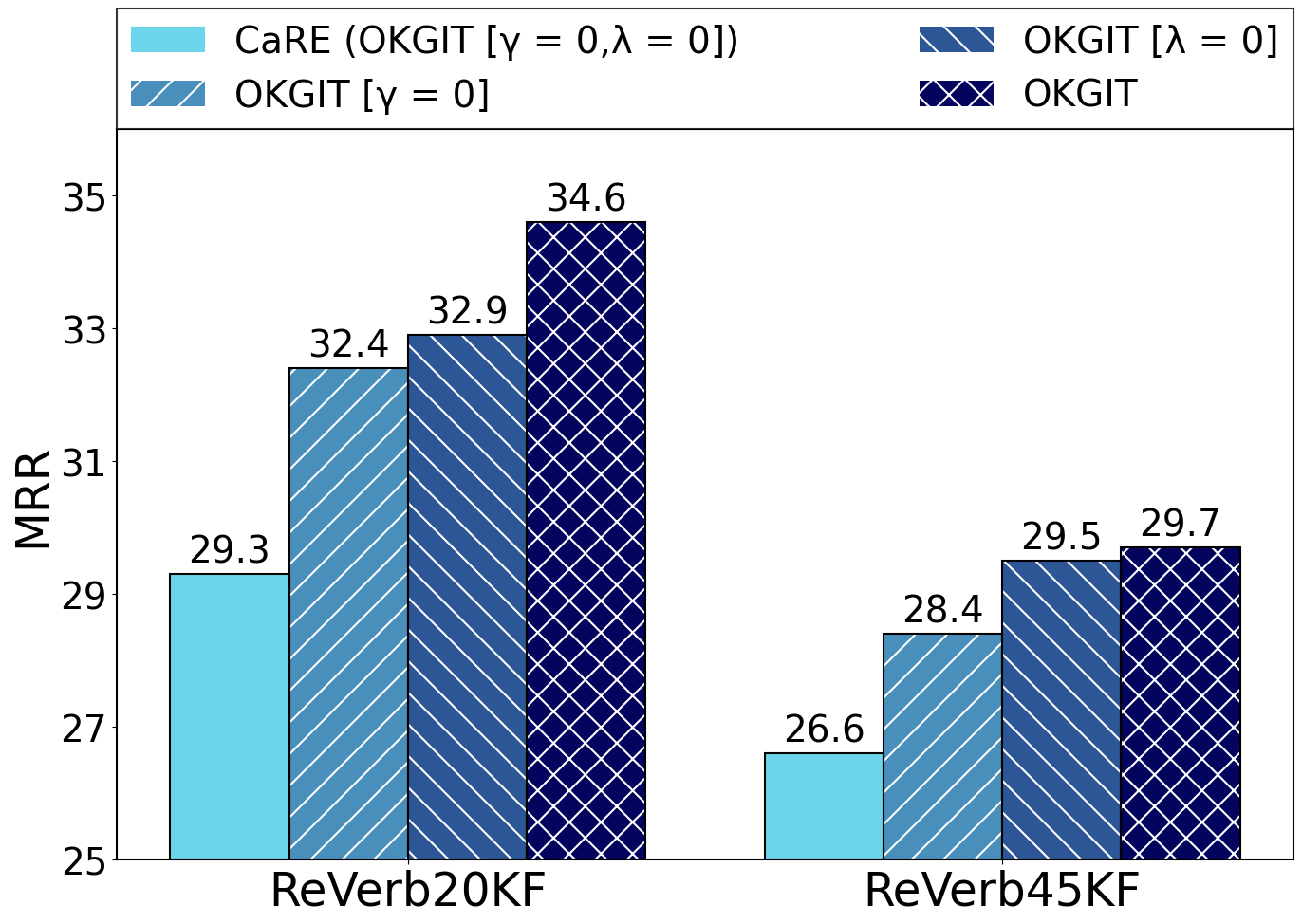}
        \caption{\label{fig:ablation_mrr}
            Effect of type compatibility score and type regularization on link prediction performance.
            While the type compatibility score with $\lambda=0$ gives better gains in MRR (11\%-12\%) than type regularization term with $\gamma=0$ (7\%-11\%), the combined model performs the best, achieving 12\%-18\% gains in MRR (\refsec{sec:lpresults}).
        }
    \end{center}
\end{figure}

The performance gain is higher for \reverbk{20} and \reverbkf{20} (+5.3 MRR) than \reverbk{45} and \reverbkf{45} (+1.2 and +3.1 MRR) datasets.
As we can see from \reftbl{tab:datasets}, the number of NPs are very close to the number of gold clusters in the 20K datasets.
Thus, the canonicalization information is slightly weaker in the 20K datasets than the 45K datasets.
Due to this, \care{} achieved better gains in the \reverbk{45} dataset as noted in \cite{ref:care}.
This leaves more scope of improvements in the 20K datasets.
By including the type information from \bert{}, \system{} is able to fill this gap.
It achieves better gains in the 20K datasets and is able to alleviate the lack of canonicalization information.
Moreover, \system{} is able to improve ranks of correct NPs ranked lower by \care{}.
This can be seen by significant improvements in the MR.

\noindent \textbf{Other Language Models}: Using \roberta{} instead of \bert{} results in similar performance improvements (\refapp{sec:app_ablations}).
However, our primary focus is to understand the impact of \emph{implicit type information} present in pre-trained MLMs, such as \bert{}, and \emph{not} to compare multiple MLMs themselves.

\noindent \textbf{Ablations}:
We perform ablation experiments to compare the relative importance of type compatibility score $\psi_{TYPE}$ and type regularization term.
We evaluate \system{} with disabled type compatibility score (\ie{} $\gamma=0$ in Equation \refeqn{eqn:scoreokget}) and disabled type regularization term (\ie{} $\lambda=0$ in Equation \refeqn{eqn:totalloss}) separately.
Please note that \care{} model is equivalent to \system{} with $\gamma=0$ and $\lambda=0$.
The results of this experiment are shown in \reffig{fig:ablation_mrr}. 
We find that while type compatibility score gives more performance gain (11\%-12\% gain in MRR) than type regularization (7\%-11\% gain in MRR), the combined model achieves the best performance (12\%-18\% gain in MRR).
It suggests that both the components are important.
Please refer to the Appendices \ref{sec:baselines}, \ref{sec:app_ablations}, \ref{sec:ufet} for more ablation experiments.

\subsection{Type Compatibility in Predicted NPs}
\label{sec:okgetTypeEval}
As noted in \cite{ref:elbert}, \bert{} vectors contain NP type information\footnote{We also verify this using Freebase, an ontological KG. Please refer to the \refapp{sec:bertTypeEval} for more details.}.
\system{} utilizes this type information for improving OpenKG link prediction.
In this section, we evaluate whether \system{} improves upon \care{} in predicting type compatible NPs. 
For such an evaluation, we require type annotations for the NPs in the OpenKGs.
However, OpenKGs do not have an underlying ontology or explicit gold NP type annotations, making a direct evaluation impossible.
Therefore, we employ a pre-trained entity typing model \ufet{} \cite{ref:ultra}.
Given a sentence and an entity mention, the entity typing model predicts the mentioned entity's types. 
Using this model, we obtain types for true as well as predicted NPs by \care{} and \system{} and use it for the evaluation.
Please note that this evaluation is limited to the coverage and quality of the \ufet{} model.

\begin{table}[t!]
\centering
\begin{adjustbox}{max width=0.49\textwidth}
\begin{tabular}{m{12em}m{5em}m{5em}}
\toprule
Dataset               & \multicolumn{1}{l}{\care{}} & \multicolumn{1}{l}{\system{}} \\
\toprule
\reverbkf{20} & 0.23                              & \textbf{0.30}                      \\
\reverbk{20}  & 0.35                              & \textbf{0.36}                      \\
\reverbkf{45} & 0.22                              & \textbf{0.31}                      \\
\reverbk{45} & 0.34                              & \textbf{0.35}                     \\ 
\bottomrule
\end{tabular}
\end{adjustbox}
\caption{\label{tab:okget_et_typeeval}
    Results of type evaluation in \care{} and \system{} predictions.
    We find that \system{} performs better than \care{} in all datasets in terms of F1-score.
    Also, the results are statistically significant for all the datasets (\refsec{sec:okgetTypeEval}).
    }
\end{table}
\begin{figure*}[thb]
        \begin{center}
        \includegraphics[width=0.40\textwidth]{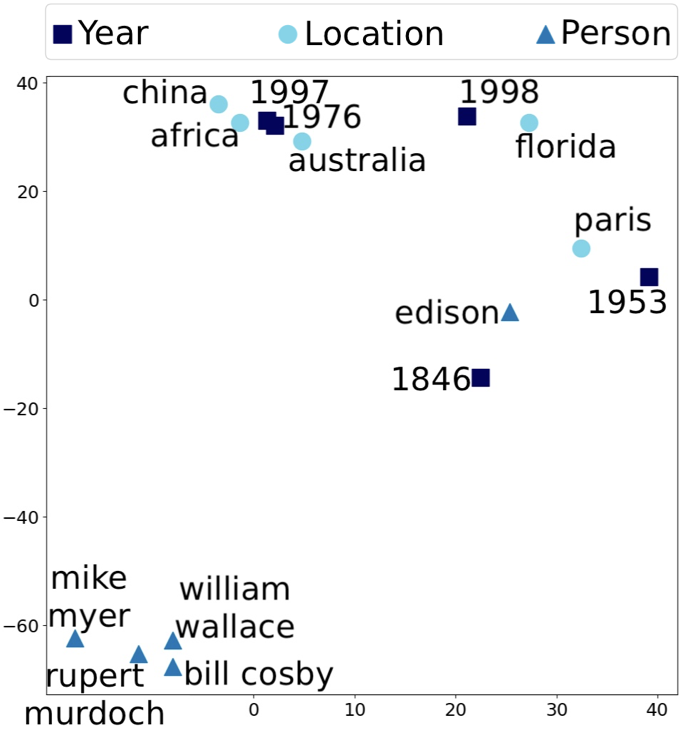} 
        \hspace{1cm}
        \includegraphics[width=0.40\textwidth]{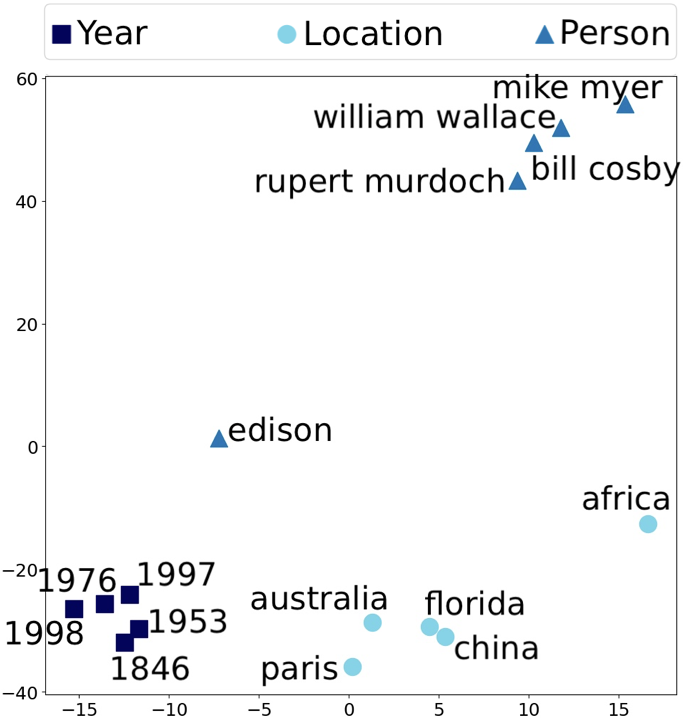}
        \caption{\label{fig:type_proj_plots} t-SNE projections of tail NP embeddings (left) and type vectors (right) extracted by the Type Projector from tail NP embeddings (\refsec{sec:type_score}) in the \reverbk{20} dataset.
            We find that the Type Projector is able to extract informative type vectors from the tail embeddings.
            This is evident from the fact that the tail embeddings corresponding to person, location, and dates were inter-mixed in the left plot,
            while they have been separated into type specific clusters in the right plot.
            Please see \refsec{sec:type_vec} for details.
        }
        \end{center}
\end{figure*}

\noindent \textbf{Evaluation Protocol:}
The type vocabulary in \ufet{} model contains $10,331$ types including $9$ general, $121$ fine-grained, and $10,201$ ultra-fine types.
The model takes a sentence $(w_1^h, \dots, w_{k_h}^h, w_1^r, \dots, w_{k_r}^r, w_1^t, \dots, w_{k_t}^t)$ formed from a triple $(h,r,t)$ along with an entity mention (either $t$ or $h$) as inputs and outputs a distribution over types.
We use the top five predicted types for our experiments\footnote{We observe similar behaviour with top one and three types.}.
For a triple $(h,r,t)$, we consider the types predicted for the true tail NP $t$ as true types $\Gamma(t)$.
Let $\hat{t}_{\care{}}$ and $\hat{t}_{\system{}}$ be the top predicted tail NP by \care{} and \system{} for the $(h,r)$ pair.
Then the types $\Gamma(\hat{t}_{\care{}})$ predicted for $\hat{t}_{\care{}}$ in the triple $(h,r,\hat{t}_{\care{}})$ is used as predicted types for \care{}.
Similarly the types $\Gamma(\hat{t}_{\system{}})$ predicted for $\hat{t}_{\system{}}$ in the triple $(h,r,\hat{t}_{\system{}})$ are used as predicted types for \system{}.
For evaluation, we calculate the mean F1-score as follows
\footnote{Since we use a fixed number of types for ground truth and predictions, precision, recall, and F1-score have the same values. Therefore, we only report the F1-score.}

\begin{equation*}
    \text{F1} = \frac{2}{n_{test}}\sum_{i=1}^{n_{test}} \frac{|\Gamma(t_i) \cap \Gamma(\hat{t_i})|}{|\Gamma(\hat{t_i})| + |\Gamma(t_i)|}.
\end{equation*}

Here, $|\Gamma(t)|$ denotes the number of types present in $\Gamma(t)$ and $\hat{t}$ represents $\hat{t}_{\care{}}$ or $\hat{t}_{\system{}}$.
We can obtain the F1-scores for head NP similarly.
We evaluate the mean F1-scores across head and tail NP prediction tasks on the test data and compare \care{} with \system{}.

As we can see from the results in \reftbl{tab:okget_et_typeeval}, \system{} performs better than \care{}, suggesting that \system{} generates more type compatible NPs than \care{} in the link prediction task.
\system{} achieves higher gains in the single-token datasets (\ie{} \reverbkf{20} and \reverbkf{45}) than multi-token dataset (\ie{} \reverbk{20} and \reverbk{45}).
Upon investigation, we found that the types obtained using the entity typing model (true as well as predicted) for the multi-tokens datasets often contain common noisy types, leading to the small difference between \care{} and \system{}.
Following \citet{ref:sigTest}, we also check the results for statistical significance using Permutation, Wilcoxon, and t-test with $\alpha=0.05$, and found it to be significant for all the datasets.

\subsection{Effectiveness of Type Projector}
\label{sec:type_vec}

To better understand the effect of type projection, we visualize the vectors in NP-space from \care{} and Type-space (\ie{} after type projection) from \system{}.
For this experiment, we randomly select 5 NPs from 3 categories, namely Person, Location and Year.
More details about this selection process can be found in the \refapp{sec:tsneSelection}.
We project the NP vectors (\ie{} $\textbf{t}$) corresponding to these NPs to a 2-dimensional NP-space using \tsne{} \cite{ref:tSNE}\footnote{We run \tsne{} for 2000 iterations with 15 perplexity.}.
Similarly, we also project the corresponding type vectors (\ie{} $\mathbf{\caretype{}}$) to 2-dimensional Type-space.
We plot the resulting vectors, color and shape coded by their respective categories, in \reffig{fig:type_proj_plots}.

We can see that the vectors from different categories in the NP-space are mixed.
However, after the type projection, the vectors in the Type-space are clustered together based on their categories.

\subsection{Qualitative Evaluations}
\label{sec:quality}
In this section, we present some examples of predictions made by \care{} and \system{} methods.
The result is shown in \reftbl{tab:examples}.
As we see in Triple-1, both \care{} and \system{} predict the correct NP (\ie{} \textit{leipzig}) on top.
However, more predictions from \system{} are type compatible (\ie{} all are locations) to the input query.
On the other hand, \care{} predictions have mixed types (\ie{} location, person, \etc{}).
Also, \care{} makes an incorrect prediction, \textit{vladimir horowitz}, possibly due to the presence of a training triple (\textit{vladimir horowitz}, \textit{had a great affinity for}, \textit{bach}).

We see similar patterns in Triple-2, where the correct tail NP should be of type \emph{number} indicating the count of votes.
\system{} is able to predict numbers in top predictions for Triple-2, while \care{} has mixed types in top predictions.

\begin{table}[t!]
    \centering
    \begin{adjustbox}{max width=0.48\textwidth}
        \begin{tabular}{m{1em}m{5.5em}m{5.5em}m{5.5em}}
    \toprule
         & Triples & \care{} & \system{}  \\ \midrule
         1. & {\begin{tabular}[c]{@{}l@{}}(\textit{bach},\\ \ \textit{moved to},\\ \ \textit{?})\end{tabular}} & 
         {\begin{tabular}[c]{@{}l@{}}\textit{leipzig}\\ \textit{mobile}\\ \textit{vladimir h.}\\ \textit{horowitz}\\ \textit{yo yo} \end{tabular}} & 
         {\begin{tabular}[c]{@{}l@{}}\textit{leipzig}\\ \textit{vienna}\\ \textit{stockholm}\\ \textit{sweden}\\ \textit{turin} \end{tabular}} \\

    \midrule

          2. & {\begin{tabular}[c]{@{}l@{}}(\textit{clinton},\\ \ \textit{lead by},\\ \ \textit{?})\end{tabular}} &
             {\begin{tabular}[c]{@{}l@{}}\textit{purchase}\\ \textit{sale}\\ \textit{video}\\ \textit{movie}\\ \textit{discount}\end{tabular}} & 
             {\begin{tabular}[c]{@{}l@{}}\textit{1500}\\ \textit{260}\\ \textit{80}\\ \textit{99}\\ \textit{hire}\end{tabular}} \\
        \bottomrule
    \end{tabular}
    \end{adjustbox}

    \caption{\label{tab:examples} Few example predictions made by \care{} and \system{} models.
        We observe that the \system{} predictions are more type compatible with the query.
        Please refer to \refsec{sec:quality} for more details.
    }
\end{table}
\section{Conclusion}

The task of link prediction for Open Knowledge Graphs (OpenKG) has been a relatively under-explored research area.
Previous work on OpenKG embeddings has primarily focussed on improving or incorporating NP canonicalization information.
While there are few methods for OpenKG link prediction, they often predict noun phrases with types incompatible with the query noun and relation phrases.
Therefore, we use implicit type information from BERT to improve OpenKG link prediction and propose \system{}.
With the help of novel type compatibility score and type regularization term, \system{} achieves significant performance improvement on the link prediction task across multiple datasets.
We also find that \system{} produces more type compatible predictions than \care{}, evaluated using an external entity typing model.

\section*{Acknowledgments}                                                                                                                                                        
We thank the anonymous reviewers for their constructive comments. This work is supported by the Ministry of Human Resource Development (Government of India).

\section*{Broader Impact}
\system{} is the first attempt towards incorporating implicit type information in OpenKG link prediction without human intervention.
It will greatly benefit densification and applications of OpenKGs where no underlying ontologies are available.

However, \system{} predictions depend on various datasets, i.e., the corpus used for training the masked language model (e.g., \bert{}) and the corpus from which the OpenKG triples were extracted.
A potential, possibly undesirable, bias may be introduced in the predictions by manipulating these corpora or adding a large number of malicious triples in the OpenKG.

We have tested \system{} in English datasets.
While the overall model architecture is independent of the language, the model's effectiveness might vary depending upon the quality of the masked language model, and it needs to be tested.

\bibliography{references}
\bibliographystyle{acl_natbib}

\appendix
\section*{Appendices}

\section{\bert{} Initialization vs Type Projectors}
\label{sec:baselines}
Here, we demonstrate the importance of type projectors by comparing \system{} with multiple \bert{}-augmented versions of \care{}.
Specifically, we initialize the phrase and word embeddings in \care{} with a pre-trained \bert{} model.
The phrase (word) is passed as input to \bert{} and the output corresponding to the $[CLS]$  token is then used for initializing phrase (word) embedding for \care{} model.
This modified \care{} model is trained similar to the base \care{} model.
Based on different initialization methods, we experiment with following baselines.

\noindent {\bf \care{} [\bert{} NP]}: NP embeddings are initialized using \bert{} and the rest of the model is same as \care{}.
        This model uses $768$ (for \bertb{}) or $1024$ (for \bertl{}) dimensional vectors.

\noindent {\bf \care{} [\bert{} NP+PROJ]}: Since \care{} [\bert{} NP] uses higher dimensional vectors ($768$ or $1024$) compared to other methods ($300$), the comparison may not be fair. To address this issue, we project \bert{} embeddings to $300$ dimension. The projection is trained with the rest of the model.

\noindent {\bf \care{} [\bert{} NP+RP]}: We initialize NP embedings as well as the word embeddings in RP encoder using \bert{} embeddings. This method also uses $768$ or $1024$ dimensional vectors\footnote{we also tried using pre-trained \bert{} as RP encoder in \care{}, however, it performed poorly due to fixed RP encoder.}.

In all the methods, including \system{}, we never fine-tune \bert{}, as our goal is to evaluate the type information already present in pre-trained \bert{} model.
We experiment with both, \bertb{} and \bertl{}, and report the best performing model.

As we can see from the results in \reftbl{tab:supResults}, \system{} outperforms these baselines.
Although \bert{} initialization improves the performance of \care{} model, the usage of explicit type-score and type regularization
leads to significant performance improvements, suggesting their importance.

\begin{table*}[thb]
\centering
\begin{adjustbox}{max width=\textwidth}
\begin{tabular}{m{12em}m{4em}m{3em}m{3em}m{3em}m{3em}m{0em}m{4em}m{3em}m{3em}m{3em}m{3em}}
\toprule
\textbf{}         & \multicolumn{5}{c}{\textbf{\reverbk{20}}}                         & & \multicolumn{5}{c}{\textbf{\reverbk{45}}}                            \\ \cmidrule{2-6} \cmidrule{8-12}
\textbf{Model}    & \textbf{MRR}($\%$)$\uparrow$ & \textbf{MR}$\downarrow$ & \multicolumn{3}{c}{\textbf{Hits}($\%$)$\uparrow$} & & \textbf{MRR}($\%$)$\uparrow$ & \textbf{MR}$\downarrow$ & \multicolumn{3}{c}{\textbf{Hits}($\%$)$\uparrow$} \\ \cmidrule{4-6} \cmidrule{10-12}
                  &             &              & {\bf @1}         & {\bf @3}        & {\bf @10}       & &              &              & {\bf @1}         & {\bf @3}        & {\bf @10}       \\ \toprule

{\care{}}                       & 30.6  & 851.1   & 24.4   & 33.1  & 41.7 & & 32.0  & 1276.8  & 25.3  & 35.0  & 44.6 \\
{\care{} [\bert{}-L NP]}        & 31.6	& 837.0	  & 24.8   & 35.0  & 44.2 & & 31.2	& 925.5   & 24.2  & 34.4  & 44.3 \\
{\care{} [\bert{} NP+PROJ]\textsuperscript{$\mathsection$}}
                                & 27.4  & 950.2   & 21.9   & 29.2  & 38.0 & & 30.7	& 952.8   & 23.0  & 34.4  & 45.5 \\
{\care{} [\bert{}-L NP+RP]}     & 30.9  & 862.4   & 24.6   & 33.5  & 42.6 & & 32.8  & 1015.6  & 25.9  & 35.9  & 45.6 \\
{\system{} [Our model]}         & \textbf{35.9} & \textbf{527.1}  & \textbf{28.2} & \textbf{39.4} & \textbf{49.9} & & \textbf{33.2} & \textbf{773.9}  & \textbf{26.1} & \textbf{36.3} & \textbf{46.4} \\
\bottomrule
\end{tabular}
\end{adjustbox}
\begin{adjustbox}{max width=\textwidth}
\begin{tabular}{m{12em}m{4em}m{3em}m{3em}m{3em}m{3em}m{0em}m{4em}m{3em}m{3em}m{3em}m{3em}}
\toprule
\textbf{} & \multicolumn{5}{c}{\textbf{\reverbkf{20}}} & & \multicolumn{5}{c}{\textbf{\reverbkf{45}}} \\ \cmidrule{2-6} \cmidrule{8-12}
\textbf{Model} & \textbf{MRR}($\%$)$\uparrow$ & \textbf{MR}$\downarrow$ & \multicolumn{3}{c}{\textbf{Hits}($\%$)$\uparrow$} & & \textbf{MRR}($\%$)$\uparrow$ & \textbf{MR}$\downarrow$ & \multicolumn{3}{c}{\textbf{Hits}($\%$)$\uparrow$} \\ \cmidrule{4-6} \cmidrule{10-12}
                  &             &              & {\bf @1}         & {\bf @3}         & {\bf @10}       & &              &              & {\bf @1}         & {\bf @3}        & {\bf @10}       \\ \toprule

{\care{}}                       & 29.3 & 308.3 & 22.1 & 31.6 & 43.2 & & 26.6 & 692.7  & 20.1 & 28.8 & 39.1 \\
{\care{} [\bert-B {}NP]}        & 31.8 & 207.6 & 24.2 & 34.8 & 46.2 & & 24.9 & 557.3  & 17.8 & 27.6 & 38.3 \\
{\care{} [\bert-L {}NP+PROJ]}   & 27.6 & 258.6 & 21.0 & 29.1 & 40.7 & & 24.7 & 600.5  & 17.4 & 27.4 & 39.2 \\
{\care{} [\bert-L {}NP+RP]}     & 30.1 & 289.3 & 22.7 & 32.8 & 43.3 & & 26.8 & 562.5  & 19.8 & 29.7 & 39.8 \\
{\system{} [Our model]}         & \textbf{34.6} & \textbf{214.7}  & \textbf{26.5} & \textbf{38.0} & \textbf{50.2} & & \textbf{29.7} & \textbf{500.2}  & \textbf{22.5} & \textbf{32.4} & \textbf{43.3} \\
\bottomrule
\end{tabular}
\end{adjustbox}
\caption{\label{tab:supResults} Results of the link prediction task.
        Here $\uparrow$ indicates higher values are better while $\downarrow$ indicates lower values are better.
        We can see that the \system{} model outperforms the baseline models on all the datasets (\refapp{sec:baselines}).
        Here, \bert{}-B and \bert{}-L denote \bertb{} and \bertl{} respectively.
        \textsuperscript{$\mathsection$}For NP+PROJ models, \bertl{} performs best for \reverbk{20}, while \bertb{} performs best for \reverbk{45}.
    }
\end{table*}
\begin{table*}[thb]
\centering

\begin{adjustbox}{max width=\textwidth}
\begin{tabular}{m{12em}m{4em}m{3em}m{3em}m{3em}m{3em}m{0em}m{4em}m{3em}m{3em}m{3em}m{3em}}
\toprule
\textbf{} & \multicolumn{5}{c}{\textbf{\reverbkf{20}}} & & \multicolumn{5}{c}{\textbf{\reverbkf{45}}} \\ \cmidrule{2-6} \cmidrule{8-12}
\textbf{Model}    & \textbf{MRR}($\%$)$\uparrow$ & \textbf{MR}$\downarrow$ & \multicolumn{3}{c}{\textbf{Hits}($\%$)$\uparrow$} & & \textbf{MRR}($\%$)$\uparrow$ & \textbf{MR}$\downarrow$ & \multicolumn{3}{c}{\textbf{Hits}($\%$)$\uparrow$} \\ \cmidrule{4-6} \cmidrule{10-12}
                  &             &              & {\bf @1}         & {\bf @3}         & {\bf @10}       & &              &              & {\bf @1}         & {\bf @3}        & {\bf @10}       \\ \toprule

{\care{} \cite{ref:care}}       & 29.3 & 308.3  & 22.1 & 31.6 & 43.2 & & 26.6 & 692.7  & 20.1 & 28.8 & 39.1 \\
{\system{}-C [$\mathbf{t_B}=[\mathbf{h};\mathbf{r}]$]} & 30.0	& 309.3	& 22.9 & 32.4 & 43.8 & & 27.1 & 666.5 & 20.2 & 29.8 & 39.9\\
{\system{}-A [$\mathbf{t_B}=\mathbf{h}+\mathbf{r}$]}   & 30.4 & 331.7 & 23.5 & 32.9 & 43.6 & & 27.1 & 660.5 & 19.9 & 30.6 & 40.2\\ 
{\system{}-R [\roberta{}]}        & 32.7  & 221.0   & 25.3   & 35.1  & 46.5  & & 29.0  & 596.7   & 21.8  & 32.0  & 43.0 \\
{\system{} [Our model]}      & \textbf{34.6} & \textbf{214.7}  & \textbf{26.5} & \textbf{38.0} & \textbf{50.2} & & \textbf{29.7} & \textbf{500.2}  & \textbf{22.5} & \textbf{32.4} & \textbf{43.3} \\
\bottomrule
\end{tabular}
\end{adjustbox}
\caption{\label{tab:app_ablations} Results of the ablation experiments.
        We replace the \bert{} module from \system{} with simple operations such as vector addition (\system{}-A) and vector concatenation (\system{}-C).
        We also use \roberta{} in place of \bert{}(\system{}-R).
        As we can see, replacing \bert{} with simple operations result in performance similar to \care{}.
        However, we do see better gains with \roberta{}, which performs better than \care{} and similar to \system{} for \reverbkf{45}.
        For all datasets, the \system{} model outperforms other variants (\refapp{sec:app_ablations}).
    }
\end{table*}

\begin{table*}[h!]
\centering

\begin{adjustbox}{max width=\textwidth}
\begin{tabular}{m{12em}m{4em}m{3em}m{3em}m{3em}m{3em}m{0em}m{4em}m{3em}m{3em}m{3em}m{3em}}
\toprule
\textbf{} & \multicolumn{5}{c}{\textbf{\reverbkf{20}}} & & \multicolumn{5}{c}{\textbf{\reverbkf{45}}} \\ \cmidrule{2-6} \cmidrule{8-12}
\textbf{Model}    & \textbf{MRR}($\%$)$\uparrow$ & \textbf{MR}$\downarrow$ & \multicolumn{3}{c}{\textbf{Hits}($\%$)$\uparrow$} & & \textbf{MRR}($\%$)$\uparrow$ & \textbf{MR}$\downarrow$ & \multicolumn{3}{c}{\textbf{Hits}($\%$)$\uparrow$} \\ \cmidrule{4-6} \cmidrule{10-12}
                  &             &              & {\bf @1}         & {\bf @3}         & {\bf @10}       & &              &              & {\bf @1}         & {\bf @3}        & {\bf @10}       \\ \toprule

{\care{}}               & 29.3 & 308.3  & 22.1 & 31.6 & 43.2 & & 26.6  & 692.7     & 20.1  & 28.8  & 39.1 \\
{\system{} (\ufet{})}   & 8.8  & 1208.2 & 6.9  & 9.6  & 11.0 & & 4.9	& 1156.8	& 1.5 & 4.0 & 11.5  \\
{\system{} [Our model]}      & \textbf{34.6} & \textbf{214.7}  & \textbf{26.5} & \textbf{38.0} & \textbf{50.2} & & \textbf{29.7} & \textbf{500.2}  & \textbf{22.5} & \textbf{32.4} & \textbf{43.3} \\

\bottomrule
\end{tabular}
\end{adjustbox}
\caption{\label{tab:ufetResults} Comparison of \system{} with \system{}(\ufet{}).
        We can see that including \ufet{} model in the system hurts the performance of the model (\refapp{sec:ufet}).
    }
\end{table*}

\section{Replacing \bert{} with other operations}
\label{sec:app_ablations}
In this section, we evaluate whether \bert{} module in \system{} can be replaced by simple operations such as vector addition and concatenation.
Specifically, we modify $\mathbf{t_B}$ in Equation (3) by replacing \bert{} with these operations leading to the following variants of \system{}.

\noindent \textbf{\system{}-C}:
\bert{} is replaced by concatenation of head NP vector $\mathbf{h}$ and relation phrase vector $\mathbf{r}$
\begin{equation*}
    \mathbf{t_B} = [\mathbf{h};\mathbf{r}].
\end{equation*}

\noindent \textbf{\system{}-A}:
\bert{} is replaced by vector addition
\begin{equation*}
    \mathbf{t_B} = \mathbf{h}+\mathbf{r}.
\end{equation*}

\noindent \textbf{\system{}-R}:
We also experiment with another masked language model \roberta{} in place of \bert{}.
\begin{equation*}
    \mathbf{t_B} = \text{\roberta{}}(h,r,\text{MASK}).
\end{equation*}

For this experiment, we use the \reverbkf{20} and \reverbkf{45} datasets as representatives.
We perform grid-search with similar hyper-parameters as in Section 5 of the main paper and select the best model based on the MRR on the validation split.
The results are reported in \reftbl{tab:app_ablations}.

As we can see from the results, the \system{}-C and \system{}-A perform very similar to \care{} on both datasets.
This suggests that the performance gains for \system{} come from the \bert{} module.
This observation is further reinforced because \system{}-R results in similar improvements compared to \care{} as \system{}.
However, in all cases, we find that \system{} with \bert{} outperforms other model variants.

\begin{table*}[thb]
\centering
\begin{adjustbox}{max width=\textwidth}
\begin{tabular}{m{12em}m{4em}m{3em}m{3em}m{3em}m{3em}m{0em}m{4em}m{3em}m{3em}m{3em}m{3em}}
\toprule
\textbf{}         & \multicolumn{5}{c}{\textbf{\reverbk{20}}}                         & & \multicolumn{5}{c}{\textbf{\reverbk{45}}}                            \\ \cmidrule{2-6} \cmidrule{8-12}
\textbf{Model}    & \textbf{MRR}($\%$)$\uparrow$ & \textbf{MR}$\downarrow$ & \multicolumn{3}{c}{\textbf{Hits}($\%$)$\uparrow$} & & \textbf{MRR}($\%$)$\uparrow$ & \textbf{MR}$\downarrow$ & \multicolumn{3}{c}{\textbf{Hits}($\%$)$\uparrow$} \\ \cmidrule{4-6} \cmidrule{10-12}
                  &             &              & {\bf @1}         & {\bf @3}        & {\bf @10}       & &              &              & {\bf @1}         & {\bf @3}        & {\bf @10}       \\ \toprule
{\care{} \cite{ref:care} }      & 30.7	& 879.2	 & 24.4	   & 33.5  & 41.7  & & 32.9	 & 1325.1  & 26.1  & 36.2  & 45.4\\ 
{\system{} [Our model]}         & \textbf{34.3}	& \textbf{609.4}	  & \textbf{27.0}   & \textbf{37.1}  & \textbf{47.4}  & & \textbf{34.1}	 & \textbf{820.2}   & \textbf{26.7}  & \textbf{37.5}  & \textbf{47.5}\\
\bottomrule
\end{tabular}
\end{adjustbox}
\begin{adjustbox}{max width=\textwidth}
\begin{tabular}{m{12em}m{4em}m{3em}m{3em}m{3em}m{3em}m{0em}m{4em}m{3em}m{3em}m{3em}m{3em}}
\toprule
\textbf{} & \multicolumn{5}{c}{\textbf{\reverbkf{20}}} & & \multicolumn{5}{c}{\textbf{\reverbkf{45}}} \\ \cmidrule{2-6} \cmidrule{8-12}
\textbf{Model}    & \textbf{MRR}($\%$)$\uparrow$ & \textbf{MR}$\downarrow$ & \multicolumn{3}{c}{\textbf{Hits}($\%$)$\uparrow$} & & \textbf{MRR}($\%$)$\uparrow$ & \textbf{MR}$\downarrow$ & \multicolumn{3}{c}{\textbf{Hits}($\%$)$\uparrow$} \\ \cmidrule{4-6} \cmidrule{10-12}
                  &             &              & {\bf @1}         & {\bf @3}         & {\bf @10}       & &              &              & {\bf @1}         & {\bf @3}        & {\bf @10}       \\ \toprule

{\care{} \cite{ref:care}}    & 28.0	& 326.0	& 21.2	& 30.5	& 41.3 & & 28.0	& 683.8	& 21.0	& 31.4	& 41.3  \\
{\system{} [Our model]}      & \textbf{31.7}	& \textbf{258.1}	& \textbf{24.2}	& \textbf{34.0}	& \textbf{46.3} & & \textbf{31.2}	& \textbf{483.3}	& \textbf{23.8}	& \textbf{34.4}	& \textbf{45.4}  \\
\bottomrule
\end{tabular}
\end{adjustbox}
\caption{\label{tab:validLPResults} Results of link prediction task on the validation split.
        We can see that the \system{} model outperforms the baseline models on all the datasets (\refapp{sec:validLPResults}).
    }
\end{table*}

\section{\care{} with Entity Typing}
\label{sec:ufet}
Entity typing is the task of predicting explicit types of an entity given a sentence and its mention.
As we are interested in improving type compatibility of predictions in the link prediction task, we can also incorporate the output from an entity typing model.
In this section, we explore this setting by replacing the \bert{} module in \system{} with an entity typing model \ufet{} from \cite{ref:ultra}. 
Specifically, we replace the vector $t_B$ in Equation (3) with the output of \ufet{} representing the predicted probability distribution over types.

\noindent \textbf{\system{}(\ufet{}) Model}: The \ufet{} model takes a sentence and an entity mention as input and produces a distribution over explicit set of types.
In our case, the sentence is formed by concatenating the subject NP, relation phrase, and object NP, while the object NP is used as mention.
The output distribution from \ufet{} is used as $t_B$ in our model. 
We call this version of the model as \system{}(\ufet{}) and compare it \care{} and \system{}.

We run a grid-search for finding the best hyper-parameter similar to Section 5 and report the results on \reverbkf{20} and \reverbkf{45} datasets.
The results are presented in the \reftbl{tab:ufetResults}.
As we can see from the results, \system{}(\ufet{}) performs poorly, even when compared to \care{}.
It suggests that explicit type vectors from \ufet{} model does not help in the link prediction task.

\section{\bert{} as Link Prediction Model}
\label{sec:bertAsLPM}
As mentioned in \refsec{sec:type_score}, $t_B$ from Equation (\refeq{eqn:bertvec}) can be used for predicting tail NPs by finding nearest neighbors in \bert{} vocabulary.
However, this approach has a limitation.
This model can only predict NPs that are single token and present in \bert{} vocabulary, restricting its applicability.

This limitation, however, is not valid for \system{}.
In \system{}, the vector $t_B$ is used for computing tail type compatibility score, instead of predicting tail NPs.
Therefore, it is not restricted to \bert{} vocabulary or single-token NPs.
As shown in \reftbl{tab:lpresults}, \system{} is equally effective for single-token datasets (\eg{} \reverbkf{20} and \reverbkf{45}) and multi-token datasets (\eg{} \reverbk{20} and \reverbk{45}).

\section{Selection of NPs for \tsne{}}
\label{sec:tsneSelection}
The OpenKGs do not have type annotations for the NPs.
Therefore, we manually annotated a set of NPs and visualized a random subset.
For this process, we first list all the NPs and shuffle them.
Then we scan this list and note the first fifteen person names, locations, and years.
Later, we select five NPs from each of these categories randomly and use them for the evaluation.

\section{Link Prediction Performance on Validation Split}
\label{sec:validLPResults}

The performance of \care{} and \system{} on validation data on the link prediction task can be found in \reftbl{tab:validLPResults}. 
These performance corresponds to the respective models which were used to report results in \reftbl{tab:lpresults} of the main paper.

\section{Type Information in \bert{} Predictions}
\label{sec:bertTypeEval}

Our proposed \system{} model is based on the hypothesis that \bert{} vectors (\ie{} $\mathbf{t_B}$ in Equation (3) in Section 4.2) contain implicit type information.
In this section, we evaluate this hypothesis that \bert{} vectors contain type information.
It should be noted that evaluating \system{} model for predicting NP types is not the goal here.
We are interested in understanding whether pre-trained \bert{} vectors have sufficient type information, measured with respect to some existing anchors.

\noindent \textbf{Evaluation Method}:
For this experiment, we use Freebase \cite{ref:freebase} which contains explicit gold type information for entities.
Specifically, we use FB15K dataset \cite{ref:transE}.
We use the data from \cite{ref:kgbert} for converting symbolic names in FB15k to textual descriptions.
We only consider the subset of triples in FB15k which has single token in the tail node as \bert{} can only predict single token NPs.\footnote{Please note that this limitation is only valid for \bert{}, not for \system{}.}
This results in $n_T=95,782$ triples.
For type information, we use the data from \cite{ref:tkrl}.
It contains 61 primary types (\eg{} \textit{/award}).
Please note that each node in FB15k can have multiple types.
For a triple $(h,r,t)$, we consider the types associated with the true tail NP $t$ as true types $\Gamma(t)$.
We then pass tokenized head NP and RP to \bert{} and find the top prediction $\hat{t} = \text{\bert{}}(h,r,\text{MASK})$ for tail position.
The set of types associated with the predicted NP $\hat{t}$, denoted by $\Gamma(\hat{t})$, is then used as the predicted types.
For evaluation, we calculate the following metrics

\begin{equation*}
    \text{Precision} = \frac{1}{n_{T}}\sum_{i=1}^{n_{T}} \frac{|\Gamma(t_i) \cap \Gamma(\hat{t_i})|}{|\Gamma(\hat{t_i})|},
\end{equation*}
\begin{equation*}
    \text{Recall} = \frac{1}{n_{T}}\sum_{i=1}^{n_{T}} \frac{|\Gamma(t_i) \cap \Gamma(\hat{t_i})|}{|\Gamma(t_i)|}, and
\end{equation*}
\begin{equation*}
    \text{F1} = \frac{2}{n_{T}}\sum_{i=1}^{n_{T}} \frac{|\Gamma(t_i) \cap \Gamma(\hat{t_i})|}{|\Gamma(\hat{t_i})| + |\Gamma(t_i)|}.
\end{equation*}

Here, $|\Gamma(t)|$ and $|\Gamma(\hat{t})|$ denotes the number of types present in $\Gamma(t)$ and $\Gamma(\hat{t})$ respectively.
\footnote{Please note that, since we have gold type annotations available for Freebase, the number of true and predicted types need not be the same. Therefore, we evaluate precision and recall along with F1-scores.}
For comparison, we use the following baseline methods to assign types to a given $(h,r,t)$.

\noindent \textbf{Random}: assign $|\Gamma(\hat{t})|$ randomly selected types.

\noindent \textbf{Most Frequent Types (MFT)}: assign $|\Gamma(\hat{t})|$ most frequent types.

\noindent \textbf{Human}: We also evaluate the type annotations provided by human annotators on randomly selected 100 triples.
    Each triple is exposed to three annotators and they are asked to provide types to the tail NP\@.
    Since most of the annotations contain one type for each triple, we take the union of the types provided by different annotators to compensate for Recall.
    For 69\% of the triples, the annotators agreed on the same type.

To be fair with the automated baselines, we use the same number of predicted types as \bert{} (\ie{} $|\Gamma(\hat{t})|$).
A comparison with a pre-trained explicit entity typing methods, such as \cite{ref:ultra}, is not applicable here as their type vocabulary is different.
As we can see from the results in \reftbl{tab:bertTypeEval}, \bert{} achieves best F1 score, suggesting that it contains type information.
The Recall for Human is low since most of the annotations contained only one type, resulting in lower F1 score.

\begin{table}[t!]
    \begin{adjustbox}{max width=0.48\textwidth}
        \centering
        \begin{tabular}{m{5em}m{5em}m{5em}m{5em}}
            \toprule
            Model  & Precision   & Recall  & F1     \\
            \midrule
            Random & 0.13          & 0.10          & 0.10          \\
            MFT & 0.45   & 0.30          & 0.31          \\
            \bert{}& 0.44 & \textbf{0.40} & \textbf{0.36} \\
            Human & \textbf{0.87}   & 0.18         & 0.27          \\
            \bottomrule
        \end{tabular}
    \end{adjustbox}
    \caption{\label{tab:bertTypeEval}
        Results of the experiment to test whether \bert{} embeddings are rich with type information.
        As we can see, \bert{} outperforms other methods in terms of F1 score, suggesting that it contains relevant type information.
        Please refer to \refapp{sec:bertTypeEval} for more details.
    }
\end{table}

\end{document}